# Generalizable Skill Learning for Construction Robots with Crowdsourced Natural Language Instructions, Composable Skills Standardization, and Large Language Model

Hongrui Yu[1], Vineet R. Kamat[2*], Carol C. Menassa[3]

[1] Assistant Professor, Department of Civil and Environmental Engineering, Virginia Tech. E-mail: hryu42@vt.edu

[2*] Professor, Department of Civil and Environmental Engineering, University of Michigan. E-mail: vkamat@umich.edu

[3] Professor, Department of Civil and Environmental Engineering, University of Michigan. E-mail: menassa@umich.edu

**Abstract**

The quasi-repetitive nature of construction work and the resulting lack of generalizability in programming construction robots presents persistent challenges to the broad adoption of robots in the construction industry. Robots cannot achieve generalist capabilities as skills learnt from one domain cannot readily transfer to another work domain or be directly used to perform a different set of tasks. Human workers have to arduously reprogram their scene-understanding, path-planning, and manipulation components to enable the robots to perform alternate work tasks. The methods presented in this paper resolve a significant proportion of such reprogramming workload by proposing a generalizable learning architecture that directly teaches robots versatile task-performance skills through crowdsourced online natural language instructions. A Large Language Model (LLM), a standardized and modularized hierarchical modeling approach, and Building Information Modeling-Robot sematic data pipeline are developed to address the multi-task skill transfer problem. The proposed skill standardization scheme and LLM-based hierarchical skill learning framework were tested with a long-horizon drywall installation experiment using a full-scale industrial robotic manipulator. The resulting robot task learning scheme achieves multi-task reprogramming with minimal effort and high quality.

**Introduction**

The construction industry has faced persistent challenges such as prolonged project durations, escalating costs, and low productivity, which are largely attributable to labor and especially skilled worker shortages (Sweis et al. 2008, Kazaz et al. 2012, Kim et al. 2020). As a result, over the past decade, the cost of procuring construction projects has increased (Stewart 2022). Construction robots are widely considered a viable solution to the labor shortage problem, given their high physical capacity and ability to manage heavy and repetitive tasks (Brosque and Fischer 2022, Cai et al. 2023, Huang et al. 2023, Ye et al. 2024, Amani and Akhavian 2024, Wang et al. 2021). However, their successful deployment hinges on the robots' ability to perform dexterous craft skills. Traditionally, these craft skills are crucial for workers in quasi repetitive work tasks where adapting various core material manipulation skills to different architectural designs, objects with similar shape but various sizes, and to address the as-built and as-designed variations is crucial for successful completion of the tasks. Largely due to the nature of such quasi-repetitive work in construction (Liang et al. 2021), programming craft skills into robots has been a significant and persistent challenge, requiring both in-depth construction domain and programming expertise.

Imitation Learning (IL) presents a conceptually promising solution to the skill transfer problem. This approach allows robots to be programmed simply by observing workers performing construction tasks, thereby improving robot programming efficiency and quality across different architectural designs and task settings. There have been several successful implementations of IL in construction robot programming (CRP), as evidenced by research from Liang et al. (2020), Huang et al. (2023), Yu et al. (2023), Wang et al. (2024), and Yu et al. (2024). In these studies, the robot observes the human doing the work and information such as, trajectory, manipulations of the material, understanding of the scene, and interactively collaborating with workers to react to the environmental observations and develop learning models.

However, the sample efficiency in IL problems has significantly hindered the adoption of such trained robots. A vast amount of time, financial cost, and physical workload for workers to provide demonstrations are needed to meet the extensive data requirements of IL problems (Johannsmeier et al. 2019, Ebert et al. 2021, Li et al. 2021, Jiang et al. 2024). Researchers explored the Hierarchical Imitation Learning (HIL) approach to address the extensive data requirements issue by decomposing a complicated computational IL model to several smaller ones and use them to imitate the 'units of action', 'basis behaviors', 'subtasks', or 'macro actions' that collectively form the whole trajectory (Schaal 1991, Physical Intelligence 2025). These "units of action" consist of the common actions a human use to perform different tasks. For example, for a construction installation task, pick up-transport-handover-manipulation will be used as units of actions (Yu et al. 2023). In addition, "elemental motions", "motion primitives" and "skill primitives" are also common terminology to represent similar concepts (Feng et al. 2013). In this paper, to match the skill transfer concept, the recently proposed term "micro skill" (An et al. 2024) was used to show the units of actions/ elemental actions. With this term defined, the process of HIL thus becomes: 1) modeling the micro skills and understanding how to finish certain motion, such as pick up or place; and 2) replicating the macro-level skill of chaining and sequencing of micro skills, in terms of inferring optimal next actions based on action histories (referred to as "macro skill" or "chaining" in this paper) (Peng et al. 2023, Zhang et al. 2023, Wang et al. 2024, Yu et al. 2024).

Another challenge in human-robot skill transfer is demonstration modality. As trajectory or vision-based action replication-based demonstration modality is prevalent, its sample efficiency is widely discussed as a challenge (Zheng et al. 2022). Vision-language-action (VLA) models still require vast amount of data even with based hierarchical models (Physical Intelligence 2025). As a result, the workload to provide sufficient demonstrations will be high. Compared to activity-based demonstrations, language-based robot skill learning has several distinct advantages, even though it has not been widely applied in CRP yet. First, language is humans' primary and most information-rich communication medium (Aceves and Evans 2024). It inherently conveys both the required elemental actions and their correct interlinking sequences, which are the two key elements in robot skill learning (Yu et al. 2024). Moreover, it is also the predominant method for human workers to provide instructions and commands in the formal and informal construction skill learning system (Sheils 1988). As a result, language-based training resources are more readily available compared to other demonstration data. Using language-based demonstrations will thus largely reduce the effort and cost of collecting demonstration data. For example, language-based demonstrations and tutorials can be easily found on various websites, such as HomeDepot or YouTube. The authors found several tutorials for diverse construction tasks, including tile installation, drywall installation, and bricklaying. The summarizations of these tutorials, in the format of micro and macro-level skills, are extracted and shown in Figure 1.

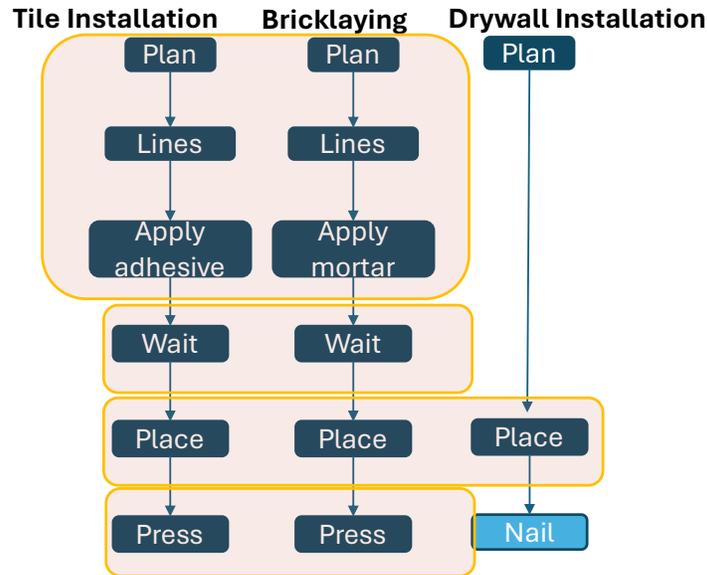

**Figure 1 Actions Required for Construction Tasks (Data Obtained from Online Tutorials –**

**Highlighted Tasks Show Similarity Among Actions Required to Complete Construction Tasks)**

The shared micro-skills illustrated in Figure 1 across the three tasks highlight the hierarchical structure inherent in verbal instructions and validate the compatibility between HIL learning and language-based IL. Humans can convey high-level skills with minimal effort through spoken instructions, making this combination highly effective in reducing demonstration effort and improving the efficiency of CRP. Traditionally, the adoption of language-based robot skill learning was hindered by the accuracy in interpreting human's language, especially given the individual instruction variability and high covariance across diverse demonstration samples and the contextual appropriateness (Francis et al. 2025) of micro skills. Sample variance can be effectively and effortlessly mitigated by incorporating diverse tutorials sourced from the Internet. For the first challenge of accuracy, the emergence of Large Language Model (LLM) successfully addressed this challenge by enabling precise extraction and processing of information from human language in various settings (Li et al. 2024). Yet, several challenges persist. First, in construction robot learning, the role and effectiveness of LLM remain unclear. Second, language-based demonstrations face difficulties in addressing high variability in personal expression. It is often challenging to correlate the action of the word with a standardized and fixed set of motions. For instance, the term "pick up", as a micro and elemental-level skill, may carry different interpretations depending on the instructor. In Pederson et al. (2016), it includes approaching the object, closing the gripper, and retracting. Conversely, Zhang et al. (2021) exclude the approach step from "pick up." Clarifying, standardizing, and disambiguating such terms, especially for sequential actions, is essential to prevent redundancy or omissions and ensure accurate motion planning for robots (Sener et al. 2022). Once micro-skills are successfully learned, the IL problem can be simplified to:

1) Adapting the micro skills with task information or based on contextual needs: such as the origin and target location for pick-and-place tasks or specific installation skills that was prototyped in Yu et al 2024. For the first one, CRP has distinct advantages compared to other robots with location information readily available from Building Information Modeling (BIM) systems.

2) Learning how to chain and interlink these elemental motions, which is an approach that is more intuitive and less computationally intensive (Wang et al. 2024, Yu et al. 2024).

Inspired by the above problems, this study proposes a hierarchical robot programming scheme that derives actions and its chaining sequences from readily available natural language instructions crowdsourced from the Internet. The proposed learning framework aims to answer these three questions: 1) What are the primary micro-level skills for construction, and how can robots be programmed to execute them? 2) How to bridge the action verbs from human verbal commands with robot motion functions without being affected by the personal variations and contextual differences? 3) How can these micro-level skills be effectively sequenced based on the specific context of different construction tasks? These questions will combine to answer the question of

6
Under review for ASCE OPEN: Multidisciplinary Journal of Civil Engineering

how to leverage the use of LLM to improve the information extraction accuracy and robot skill learning efficiency.

The authors address these questions through extensive literature review and computational model building with a bottom-up approach. The steps are outlined as follows:

1. Building a standardized database of elemental activities to eliminate personal variations in language-based instructions. First, an extensive literature review is performed to understand what common elemental motions are mentioned in assembly, manufacturing, and other robot manipulation research. Observing that action words standardization and unified definitions were missing from the current elemental motion studies, a framework to refine, coalesce, and systematize the meaning of different elemental motions to form a robot elemental motion dictionary is developed. The details are shown in Section 3. In Pastor et al. (2009), this process is referred to as "attaching sematic".

2. Computing the skill chaining process: The work presented in this paper also develops an efficient computing scheme that retrieves skills from natural language instructions to connect and chain the elemental motions. Several computational models, such as Markovian models, network-based frameworks, and language models, were evaluated to determine the most suitable option for this learning task.

3. Prototype the BIM semantic information communication pipeline for micro skill parameterization: The authors propose a data pipeline that retrieves task information directly from the BIM model and sends the coordinates directly to the robot programming system to finish lower-level learning for parameterizable elemental motions. An engineering system was built to validate the feasibility of this idea.

**2. Literature Review**

This section provides a literature review of the current research in LLM and robot learning. Micro-level motions and chaining learning (which refers to the process of learning the correct sequence of actions) were also systematically studied with the commonly used actions summarized.

*2.1 LLM and Hierarchical Skill Learning*

Language-based skill learning improves skill discovery and acquisition efficiency and scalability (Redis et al. 2024, Ha et al. 2023). Several studies have validated its effectiveness in reducing the human demonstration workload for diverse general robot motion and skill learning tasks, such as walking and squatting (Chen et al. 2024, Redis et al. 2024).

LLM has diverse functions and roles in robot learning. The combination of LLM and hierarchical skill learning or zero-shot learning has recently gained significant attention (Peng et al. 2023, Zhang et al. 2023, Ha et al. 2023, Redis et al. 2024). In these combinations, first, LLM can be used as the task decomposer and planner. For example, Ha et al. (2023) combined LLM, visual models, and diffusion policy generation models to transfer multi tasks. LLM serves as the task decomposer and hierarchical planner in the beginning of the whole skill interpretation and learning process. Its zero-shot learning capability was commonly leveraged by generating the success-detection functions for each task and inferring success conditions. In Redis et al. (2024), LLM plays a similar role to plan generator or motion planner with several auxiliary functions added to improve the quality of LLM usage. Redis et al. (2024) also added a rephrasing function to unify the language (English, Danish, and French) data before importing them to the LLM module. A conformance checking/ similarity measures was also added after the LLM module for quality control.

LLM is also widely used in the skill chaining process of hierarchical skill learning. In Zhang et al. (2023), LLM was used to bootstrap acquired skills and adopt new long-horizon behaviors online for water pouring tasks in domestic settings. The authors use LLM to infer the most likely next action considering action histories and perform action chaining. It improves the exploration Reinforcement Learning (RL) efficiency and promotes continuous growing of robot skill libraries with lower learning costs. The zero-shot reasoning capability and hierarchical skill learning advantages of LLM are combined in Peng et al. (2023). In this work, LLM was used reversely to decompose the tasks to subtasks to gap the sematic knowledge in language instructions with the environment observations for self-driving applications. It was used as the interface between human experts and robots. In construction automation research, there are limited studies on hierarchical reinforcement learning or hierarchical LLM-based methods.

Peng et al. (2023) raised a critical question of how to breakthrough the information limitation in the language instructions of human experts. In human society, language forms a series of coordinate changes and represents a common pattern of motion. For example, "place" means to drop off things to a location from the language-only perspective. However, in several verbal or video construction tutorials, it also implies to release the object. In this case, the "release" is an unknown and hidden information to LLM. Based on this observation, this paper proposes that robot micro skills should be explicitly defined to align human's verbal instructions with the implicit and unspoken actions expected in the human mind.

## *2.2 Micro Skill Database for Hierarchical Imitation Learning*

Extensive discussions have focused on defining micro-skills and developing a standardized skill database to support HIL. Liu et al. (2022) mentioned the skill database should include action library, task library, and scene library. The proposed skill database was validated through a pick-and-place case study. In Miao et al. (2022), this skill base would not be complete if without task, agent, motion, and object information (for catering tasks). Nozaki et al. (2014) proposed that sensory input should also be included. They divided the haptic information into transformation matrices, hybrid angles, pure position commands, and pure force commands for human hand motion imitation tasks. Micro vibration analysis for identifying grasping and manipulating actions was included. Nonetheless, the name of the actions and their definitions in terms of robot planning are commonly mentioned in micro skill knowledge base building.

In the above studies, one key metric for evaluating the quality of proposed micro skill sets is continuity. Micro skills need to be combined to be able to represent a full chain of actions. The discontinuity bound has been widely discussed. For IL, the discontinuity in the demonstration could reduce the number of viable analytic solutions in robot motion planning and execution of learned tasks (Hönig et al. 2022, Ortiz-Haro et al. 2024).

A more extensive literature review was conducted by the authors to define construction manipulation-suitable micro-skills and explore the implications of a continuous definition. Several keywords search queries for scholarly articles were performed, including knowledge base, motion primitives, robot skill representations, and hierarchical knowledge representation. The example combined search queries are shown as follows: "robot" + "knowledge graph" + "assembly", "robot" + "motion primitives" + "assemble", "hierarchical knowledge representation dynamic motion primitives", "robot skills for manufacturing", "robot" + "motion primitives" +"knowledge graph", "robot" + "unit of actions ", and "robot" + "skill representations". However, not all papers explicitly mention the name of micro skills.

Eventually, 86 papers were selected from the above search results based on relevancy and 35 sets of elemental motions were summarized, as shown in Table 1. The mismatch in the number of relevant papers and the numbers of available action dictionaries reflects the popularity and sufficiency (for fields other than construction) of non-transparent robot learning models (i.e., not understanding what the state or action variables explicitly represent). The summary of the domains or task scenarios of the 20 studies are shown in Figure 2.

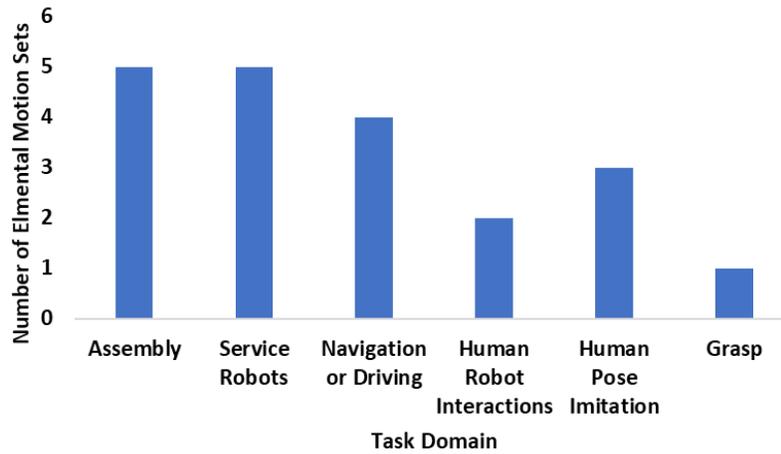

**Figure 2 Summarization of Elemental Motion Sets by Work Domain or Action Area**

Figure 2 illustrates that a substantial portion of the research focuses on assembly papers, which conceptually have a significant overlap with construction site activities (except for the quasi-repetitive nature of the latter). The relevance of these domains is underscored by utilizing elemental motions from assembly contexts to guide the subsequent analysis and modeling efforts.

Table 1 Sample Summary of Micro Skill Knowledge Bases in Existing Studies

| Paper Citation | Actions | Tasks |
|---|---|---|
| Agostini et al. 2020 | Reach and pre-grasp, reach and grasp, grasp, lift | Domestic, pouring task |
| Tenorth et al. 2012 | Reach to approach, open gripper for grasp, reach, close gripper, lift object to approach, reach to stop | Grasp bottle |
|  | Move to grasp, grasp, move to handover, handover, open gripper | Serve drink |
| Mayr 2024 | Pre-, hold, post-manipulation | Pick and place/general |
| Niekum et al. 2025 | Start, reach toward, pick up, return to home pose, place, return to home pose | Pick and place |
|  | Start, left 1, left 2,… right 1, right 2 | Whiteboard survey |
| Tenorth and Beetz 2017 | Perceive (features objectDetected), reach, grasp (features touchEvent), lift (features collisionDetected) | Pick up |
| Denk, and Schmidt 2001 | Pre-swing, swing, heel-contact | Walk |
| Stenmark and Malec 2015 | Move, convey, grasp, release, wait | Manipulation |
| Forte et al. 2012 | Reaching and grasping | Manipulation |

| Sener et al. 2012 | Attach, track, screw, nail, pick up, remove, put down, position, push | Assembly |
|---|---|---|
| Zhang et al. 2021 | Transition, pick up, install, search | Assembly |
| Pederson et al. 2016 | Pick up, equals to move + move relative + close gripper + move relative | Manufacturing |
| Chen et al. 2022 | Carry, hold, attach, release, transit, transfer, hold | Assembly |
| Lee et al. 2024 | Approach, grasp, move, hold, place, insert, screw, rotate, slide | Assembly |

The full action verbs list can be seen in Appendix 1. As can be seen in Table 1, micro skills have different meanings in different task contexts and with different human experts' preferences. However, some common skills are frequently observed, including pick up, place, install, attach, move, grasp, especially for manipulations or assembly tasks. In addition, the above papers also mention the corresponding robot action planning functions and success metrics for the given skills. For example, in Lee et al. (2024) "Human-Robot Shared Assembly Taxonomy: A step toward seamless human-robot knowledge transfer", place means "the process of placing the manipulated object on the surface of the target object at a certain pose". The corresponding robot motion planning function can set the destination to be the surface of the target object. Nonetheless, certain pose remains vague with the above definitions. To address this problem, in this paper, we propose to further define the skills and ease the robot motion planning by parameterizing the skills with tasks and object geometric configurations from Building Information Modeling (BIM).

## 3. Methodology

### *3.1 Overview of Methodology*

The Literature Review section emphasizes that hierarchical skill learning combined with LLM significantly improves robot programming efficiency. By dividing large-scale training problems into smaller, manageable models, this approach streamlines the learning process. On this basis,

LLM enhances efficiency further by extracting information from natural language instructions and facilitating the direct transfer of human expertise and knowledge to robots.

Combining these two advances achieves a parallelized and more effective framework for robot learning and skill transfer. However, it is unclear how to build a computational framework to include both advances and what are elemental motions that form the long-horizon and large-scale learning problems. To address these questions, this paper proposes a HIL approach with LLM learns higher-level control policies (macro learning) and forms a comprehensive database to cover common elemental construction manipulation skills (micro learning). This section is organized around hierarchical skill modeling, with two levels in the bottom-up way: micro-skill learning (Section 3.1) and macro-skill learning (Section 3.2). Each section highlights distinct roles of LLM, integrating innovations from Peng et al. (2023) and Redis et al. (2024). An overview of the methodology structure is presented in Figure 3. The engineering system prototyped to validate the feasibility of the proposed robot learning architecture was also demonstrated, shown in Figure 4.

*3.2 Standardized Micro Skill Learning with Task Parameterization*

The hierarchical modeling process begins with lower-level micro-skill modeling. The primary objective of this stage is to collect, unify, and parameterize elemental actions in construction activities. To address these aspects, the elemental motion knowledge base identified in the Literature Review section is collected and refined by focusing on resolving a few observed issues (such as unclear definitions for the same action verb) and perform a granularity matching process, as detailed in Section 3.2.1. In addition, human language instruction patterns derived from two large-scale human-robot interaction databases are examined to align the micro skill database to the human's verbal instructional habits (Section 3.2.2). Additionally, an information exchange pipeline from BIM to Robot Operating System (ROS) is established to facilitate the parameterization of micro-skills in relation to specific tasks. More details about the BIM-ROS framework can be seen in section 3.2.3.

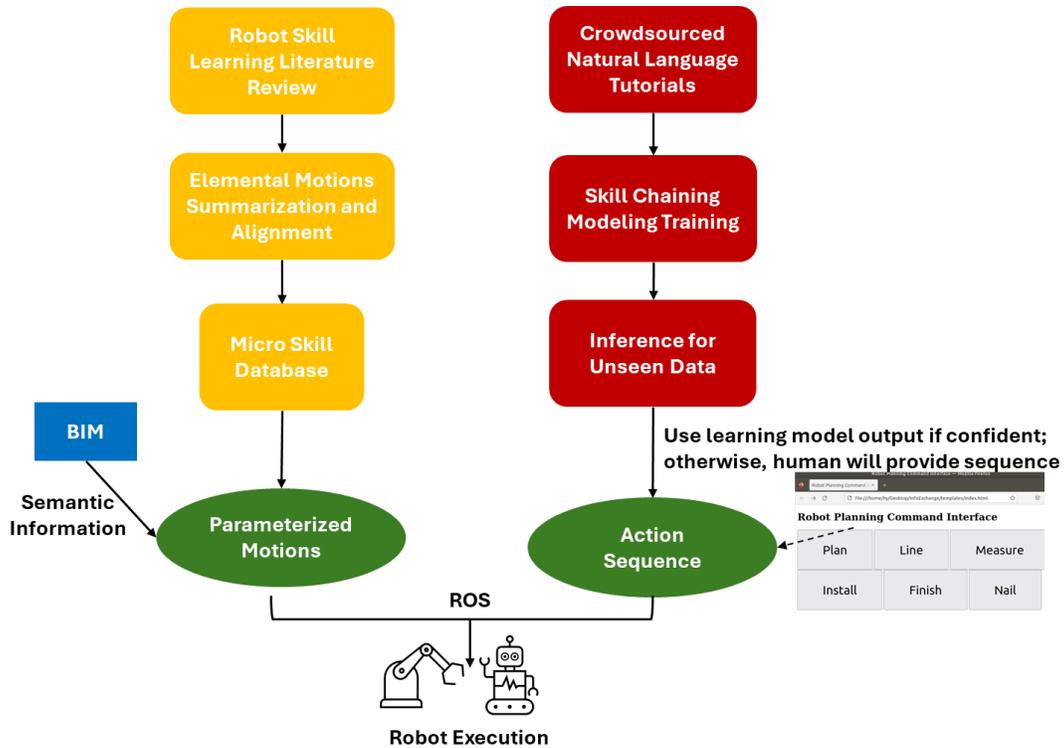

Figure 3 Overview of Research Methodology

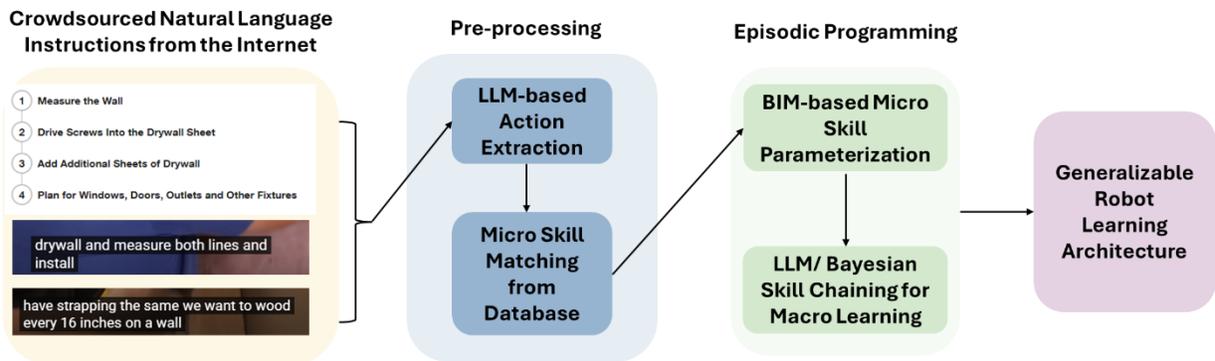

Figure 4 Proposed Generalizable Robot Programming Architecture (Source: HomeDepot.com and YouTube.com)

*3.2.1 Construction Robot Micro Skill Database*

First, as listed in Table 1, an extensive literature review was performed to explore as many robot micro skill knowledge bases in the existing studies. The comprehensive list is shown in Appendix 1 Table 1. The authors also conducted a micro skill conformance check and validated the suitability of the micro skills from assembly studies. This was performed by leveraging state-of-the-art LLM. Crowdsourced construction tutorials (mostly tile installations and bricklaying) are processed to extract the elemental motions or activities for each step. To standardize the output and improve robot learning data quality, the authors refined prompts templates by adding human's task instruction, role definition, perception Application Programming Interface (API) descriptions, and explanation of the instruction following task space to the decomposition prompts. The raw elemental action verbs extracted are "align", "attach", "carry", "close gripper", "grasp", "handover", "hold", "insert", "install", "move", "move relative", "pick up", "position", "push", "put down", "release", "remove", "screw", "search", "track", "transfer", "transit", and "transition". Through this step, the alignment between assembly micro skills and construction skills was confirmed.

In addition, to improve the quality of the definition of elemental actions, several rules were added.

1) The micro skills in this dataset should cover all the motions needed for the given construction activity, meaning there should be no crucial motion missed when representing a construction task with a set of micro skills.

2) The micro skills in this dataset should be mutually exclusive, meaning there should not be repeated motions between any two micro skills.

3) The micro skills should have at least one preceding or successor event (including "start" and "finish") to ensure the continuity in the state transformation of manipulated objects.

The above rules can be represented with the following mathematical equations:

With given micro skill dataset $M = \{s_1, s_2, \ldots, s_n\}$, which contains a total number of $n$ micro skills: $s_1, s_2, \ldots, s_n$

1) Given task $t_i = \{s \mid s \in M \text{ and } s \text{ is micro skill needed for this task}\}$ $t_i \subseteq M$ or $t_i \in P(M)$, where $P(M)$ is the power set of $M$;

2) For two distinct micro skills $s_i, s_j \in M: s_i \cap s_j = \phi$, for all $i \neq j$;

3) Given any micro skill and its associated manipulated object states: $s_i.O_{start} = $ object start state of skill $s_i$ and $s_i.O_{end} = $ object end state of skill $s_i$, $\exists s_j \in M, i \neq j: (s_i.O_{start} = s_j.O_{end}) \vee (s_i.O_{end} = s_j.O_{start})$;

The necessity of abovementioned rules arose upon scrutinizing the action verbs and observing their characteristics, including:

- Redundancy in meaning: Some verbs are synonymous. For example, handover could mean the transition process only, or it could include the grasping actions before the object transition. To address this, verbs have been reclassified when input to the micro skill database, based on the anticipated final state of the manipulated object upon successful completion of the action. Actions that result in similar object end states are consolidated under the synonym category. Conversely, actions leading to distinct outcomes are allocated unique entries.

- Divergence in planning complexity: The action verbs exhibit varying degrees of planning intricacy. While some denote simple, single-step motions (e.g., "push"), others encapsulate more complex or composite movements (handover = pick up + object transition). The concept of granularity matching concept in data analytics and natural language process to select only single-step motions. More details will be provided in the Discussions and Limitations section.

- Continuity and comprehensive activity representation: All micro skills should collectively include all possible actions required for the completion of a given construction task to ensure no gaps or discontinuities arise in the manipulated object states. To ensure continuity, definitions of micro skills are provided with the initial and terminal object states associated with each subtask. Continuity in the action chain is maintained by ensuring that the initial object state for any given action aligns seamlessly with the object state at the conclusion of the preceding subtask, as shown in Table 2.

Table 2 Sample Action Verbs from Robot Activity Database

| Word | ObjectStateBegin | ObjectStateEnd | Synonyms |
|---|---|---|---|
| Install | At least 0.2 meters (Wang et al. 2024) away from the target location | Coordinate of object center same as the target center | Connect, Position |
| Align | The orientation of the object is different from the target or at least one of the three linear coordinates of (x,y,z) is different from that of the target | The orientation of the object is same as the target and at least two of the three linear coordinates of (x,y,z) is different from that of the target | Not Applicable |
| Pick up | Object coordinate (center of object) different from gripper center but its z coordinate will be same as one of object in the scene (holder) | Object coordinate (center of object) is same as the gripper center | Not Applicable |
| Transfer (Specifically refers to giving an object to another human or robot. Otherwise referred to as Grasp) | Object coordinate (center of object) is same as the gripper center | Object coordinate (center of object) is same as the human hand center | Transition, Handover |

*3.2.2 Human Instruction Behavior Analysis*

The analysis and derived rules establish the core definitions of micro-skills. However, it remains uncertain whether these definitions align with humans' habitual instructions. Such alignment is critical to ensure that robots correctly interpret and execute tasks based on natural language commands. To address this, a data-driven approach is employed by analyzing human-instructed actions and their associated object final states using two large-scale benchmark robotics datasets with natural language instructions: TEACH (Padmakumar et al. 2022) and ALFRED (Shridhar et al. 2020). The ALFRED dataset contains 25,743 directives and 8,055 expert demonstrations. It has seven actions: Pickup, Put, Open, Close, ToggleOn, ToggleOff, and Slice.

For the TEACH dataset, it has the object trajectory for each action and each file contains all actions and trajectories required to finish the task. The actions include: Stop, Move to (code: 2), Forward (code: 4), Backward (code: 5), Turn Left, Turn Right, Look Up, Look Down, Pan Left, Pan Right, Move Up, Move Down, Double Forward, Double Backward, Navigation, Pickup, Place, Open, Close, ToggleOn, ToggleOff, Slice, Dirty, Clean, Fill, Empty, Pour, Break, BehindAboveOn, BehindAboveOff, OpenProgressCheck, SelectOid, SearchObject, Text, Speech, and Beep. The final object state associated with each action is extracted from the middle-size Execution from Dialog History (EDH) training dataset. 181,407 entries were obtained with our preprocessing scripts, and some samples can be seen in Figure 5.

These two datasets supported the validity of the micro skills included in the proposed skill base in several ways. First, the actions commonly selected by human experts to command robots in these datasets closely align with those defined in the micro-skill base, indicating consistency with natural human instructional preferences. Second, the trajectories associated with these actions were manually reviewed to interpret their meanings. For instance, the action "forward" does not denote a specific direction, such as toward the positive x-axis, as no consistent patterns were observed. This observation aligns with the proposed definition of move forward in the knowledge base. Third, the granularity of the data was analyzed to clarify the contextual meanings of actions. For example, while "move to" and "forward" may carry identical meanings in some contexts. When they do, it was hypothesized that "move to" and "forward" would not occur sequentially in an action sequence. To test this hypothesis, Bayesian transition probabilities between actions (e.g., transitions from action 2 to 4 versus 2 to 5) were calculated. The results, displayed as a heatmap in Figure 6, support this hypothesis.

| | A | B | C | D |
|---|---|---|---|---|
| 1 | action_id | pose | file | action_change |
| 2 | 2 | [1.75, -2.75, 0.9009991884231567, 0, 30.000024795532227, -270.0] | 0008f3c95e006303_2053.edh0.json | TRUE |
| 3 | 4 | [1.75, -1.0, 0.9009991884231567, 0, 30.000028610229492, -180.0] | 0008f3c95e006303_2053.edh0.json | TRUE |
| 4 | 2 | [1.5, -1.0, 0.9009991884231567, 0, 30.000028610229492, -180.0] | 0008f3c95e006303_2053.edh0.json | TRUE |
| 5 | 5 | [0.0, -1.0, 0.9009991884231567, 0, 30.000024795532227, -270.0] | 0008f3c95e006303_2053.edh0.json | TRUE |
| 6 | 8 | [0.0, -0.75, 0.9009991884231567, 0, 30.000028610229492, -0.0] | 0008f3c95e006303_2053.edh0.json | TRUE |
| 7 | 2 | [0.25, 1.0, 0.9009991884231567, 0, 30.000028610229492, -0.0] | 0008f3c95e006303_2053.edh0.json | TRUE |
| 8 | 4 | [1.5, 1.0, 0.9009991884231567, 0, 30.000024795532227, -270.0] | 0008f3c95e006303_2053.edh0.json | TRUE |
| 9 | 2 | [1.5, 1.25, 0.9009991884231567, 0, 30.000024795532227, -270.0] | 0008f3c95e006303_2053.edh0.json | TRUE |
| 10 | 4 | [1.5, 2.75, 0.9009991884231567, 0, 30.000028610229492, -180.0] | 0008f3c95e006303_2053.edh0.json | TRUE |
| 11 | 2 | [1.25, 2.75, 0.9009991884231567, 0, 30.000028610229492, -180.0] | 0008f3c95e006303_2053.edh0.json | TRUE |
| 12 | 4 | [0.5, 2.75, 0.9009991884231567, 0, 30.000024795532227, -90.0] | 0008f3c95e006303_2053.edh0.json | TRUE |
| 13 | 8 | [0.75, 2.75, 0.9009991884231567, 0, 30.000024795532227, -90.0] | 0008f3c95e006303_2053.edh0.json | TRUE |
| 14 | 2 | [1.25, 2.5, 0.9009991884231567, 0, 30.000024795532227, -90.0] | 0008f3c95e006303_2053.edh0.json | TRUE |
| 15 | 5 | [1.25, 0.5, 0.9009991884231567, 0, 30.000028610229492, -180.0] | 0008f3c95e006303_2053.edh0.json | TRUE |
| 16 | 4 | [1.25, 0.5, 0.9009991884231567, 0, 30.000028610229492, -180.0] | 0008f3c95e006303_2053.edh0.json | TRUE |
| 17 | 2 | [1.0, 0.5, 0.9009991884231567, 0, 30.000028610229492, -180.0] | 0008f3c95e006303_2053.edh0.json | TRUE |
| 18 | 2 | [1.75, -2.75, 0.9009991884231567, 0, 30.000024795532227, -270.0] | 0008f3c95e006303_2053.edh1.json | TRUE |
| 19 | 4 | [1.75, -1.0, 0.9009991884231567, 0, 30.000028610229492, -180.0] | 0008f3c95e006303_2053.edh1.json | TRUE |
| 20 | 2 | [1.5, -1.0, 0.9009991884231567, 0, 30.000028610229492, -180.0] | 0008f3c95e006303_2053.edh1.json | TRUE |
| 21 | 5 | [0.0, -1.0, 0.9009991884231567, 0, 30.000024795532227, -270.0] | 0008f3c95e006303_2053.edh1.json | TRUE |
| 22 | 8 | [0.0, -0.75, 0.9009991884231567, 0, 30.000028610229492, -0.0] | 0008f3c95e006303_2053.edh1.json | TRUE |

Figure 5 Screenshot of TEACH EDH Dataset Final State Coordinates of Actions

The analysis of instructions in ALFRED followed a slightly modified approach. To enhance efficiency and ensure consistency with the micro-skill granularity outlined in Section 3.1.1, only the "pick up" and "put" actions were extracted and examined. The ALFRED training dataset includes 890 "pick up" actions and 954 "put" actions. An unexpected pattern emerged: all coordinates associated with these actions—both translational and rotational—were recorded as zero. This indicates that the "pick up" pose is entirely determined by the object's state, requiring the robot's pose to align with the object's position. Similarly, the end pose of "put" must match the target's state. These findings are consistent with the micro-skill definitions presented in Table 2, which means that the micro skills defined in Table 2 conform with the human's natural instructional habits.

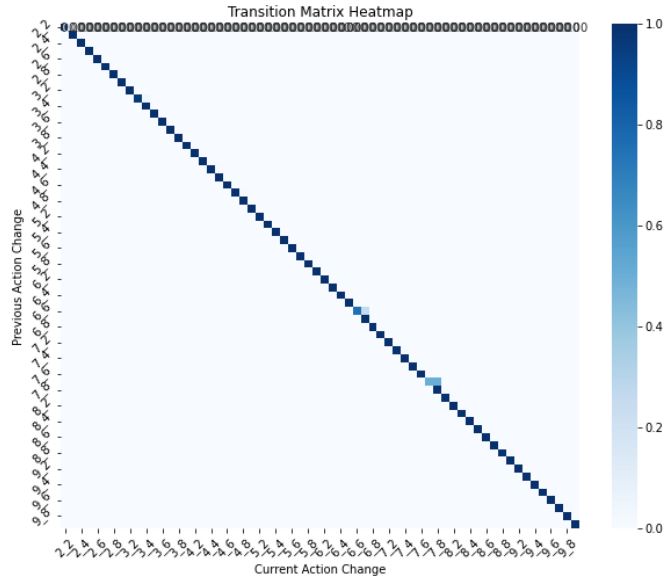

Figure6 Transition Probability of EDH Motions

3.2.3 HRC Modes Illustrations

To complete micro skill parameterization process, the authors built a BIM–Web App– ROS information system, as shown in Figure 7.

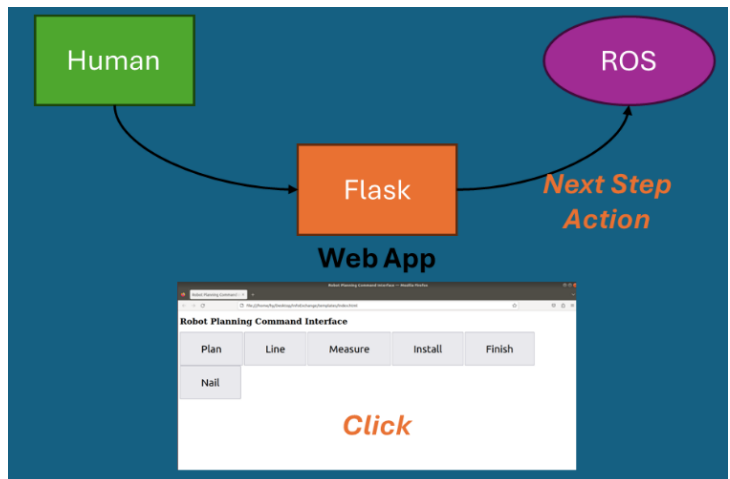

Figure 7 Information Flow in the Human-Supervised Execution Process

This system passes material and target locations in 3D models in Rhino directly to a web application supported by the flask library, which enables lightweight web application writing with Python. Humans choose micro skills by clicking the skills in the correct sequence. The skills will be sent to ROS to trigger the execution of programmed micro skills via the flask_ros library. The robot will execute the corresponding motions based on commands from the web application.

*3.3 Skill Chaining with Macro-Level Learning*

The repetitive clicking and selecting micro-skills mode shown in Figure 7 can increase workers' mental workload and lead to fatigue. This section aims to automate the micro-skill chaining process using machine learning models. To mitigate the effects of limited human supervisors' proficiency on the robot's learning performance and enhance the quality of input data, a pipeline is proposed to extract relevant information from crowdsourced, natural language-based construction tutorials available online. Additionally, this paper experimentally evaluates several approaches, generating new insights into the effectiveness of different methods for addressing the macro level learning problem.

*3.3.1 Crowdsourced Natural Language Instructions Collection and Preprocessing*

While individual differences in demonstrations exist, the impact of these variations can be mitigated by sourcing multiple demonstrators with diverse skill levels and preferences. Instead of relying solely on workers from specific regions—which may introduce regional biases or incur high costs—this study suggests using demonstration data available on the web. The primary sources of this data are YouTube videos and written tutorials from trade and home improvement stores' websites. To ensure the quality of the demonstration data, several standards were enforced:

- Select Highly Viewed Videos: While high viewership does not necessarily equate to quality, it often indicates viewer trust and perceived quality. Choosing videos with a large number of views can be a practical proxy for their reliability and educational value. The authors also randomly sampled a few videos to check the accuracy of information and quality of construction skill teaching.

- Opt for Reputable Sources: Prefer videos from websites that not only have high viewership but are also endorsed by industry leaders, such as Home Depot or Armstrong Ceilings. This approach leverages the credibility and established trust of such platforms.

- Standardize Data Modality: Despite the growing interest in multimodal data, this paper focuses on standardizing the data modality by exclusively utilizing text. This includes text tutorials and captions from YouTube and other videos. This decision simplifies the demands on the learning and data processing models. Using a single modality is likely to enhance the performance of the chosen learning model, as it reduces complexity and focuses the learning process.

For video tutorials, the text transcript (English-only for this round of data collection) of the video was extracted using either the GlaspAI Chrome extension or the YoutubeTranscriptAPI library. The transcripts of both the article-based and video-based tutorials were then manually reviewed and preprocessed to remove any irrelevant information (e.g., advertisements) or to fix any spelling errors created in the video to text transcript translation.

The data collection started by selecting the five most common construction tasks: drywall installation, ceiling installation, tile installation, bricklaying, and window installation. For each task, we gathered ten tutorials: five from various websites and five from YouTube videos.

The processed transcripts were then input into ChatGPT to extract key action words from each tutorial. Several experimentations with prompt-engineering were conducted to determine a prompt that produced the most accurate and well-formatted action verb list. Given the relatively small size of the dataset, we employed a qualitative validation approach to assess the accuracy of ChatGPT's action extraction. Specifically, the authors evaluated outputs based on two criteria: (1) comprehensiveness—whether all key steps from the original tutorial were included in the output, and (2) sequence accuracy—whether the extracted actions followed the correct order as presented in the original tutorial. Prompts were only adopted if both criteria were satisfied. To ensure robustness, we tested the prompt on a set of five tutorials selected from diverse sources, including different websites and a mix of video scripts and purely textual tutorials. Only after confirming consistent performance across these examples did we proceed to apply the method to a larger dataset. After receiving this prompt, ChatGPT often returned its list of action verbs with verbs delimited by certain characters such as a vertical line (|), commas, or back-slashes. To further extract each individual action verb, a Python script was developed to split up an action verb list by a specific delimiter and extract the individual action verbs into an array. This array of action verbs was then used to modify and customize the micro skills displayed on the web application. The whole preprocessing step information process can be seen in Figure 8.

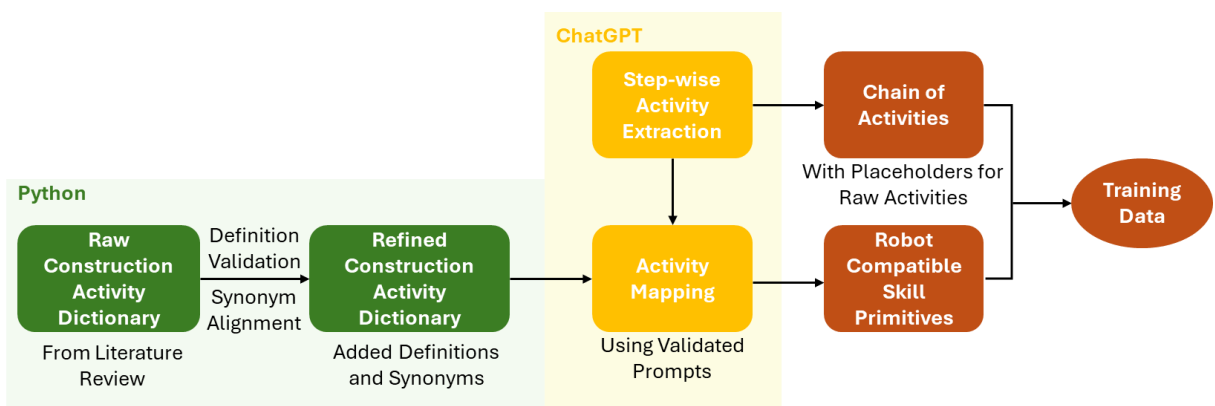

**Figure 8 Information Flow for Preprocessing with ChatGPT**

The next step in the process was to extract elemental motions and align them with an action verb dictionary. This task commenced with manually preprocessing the data using Python to isolate step-by-step instructions. Subsequently, we leveraged the advanced capabilities of the natural language processing model, ChatGPT. The authors observed several errors, including hallucinations, replicated synonyms, and inaccurate sequencing. To solve these issues, several error-handling experiments were performed with the sample set of 5 tutorials, as shown in Table 3. For each experiment, the authors tested and evaluated several prompts to determine their effectiveness in extracting action verbs from the step-by-step instructions. The most suitable one is "Process the following text by breaking it down into only action words. Map the instructions in the text to action words provided and display one action word per numbered step that summarizes that step. These action words should only be from the action word list provided. If you cannot map a given step to an action word found in the list, try to map synonyms of the instruction to an action word from the list. Either way, do not display any action words that are not in the action word list provided. Only assign one action word per step. Do not display headings for each step, only the action word. Here are the action words you can map to: [] Here is the text to process: []". This prompt achieved the optimal performance due to the level of detail provided to the task.

Table 3 Performance Guardrail Processes and Metrics

| Error Handling Experiments and Measures | Method | Performance Metrics | Output and Solutions |
|---|---|---|---|
| Action Extraction Prompt Engineering | Try different prompting methods, including chain-of-thoughts prompts to | **1) Comprehensiveness:** all key steps present | Optimal chain-of-thought prompt to use for the action extraction step: *"Process the following text by breaking it down into only action words. Map the instructions in the* |
| | provide more specific requirements | **2) Sequence accuracy:** actions follow original order | *text to action words provided and display one action word per numbered step that summarizes that step. These action words should only be from the action word list provided. If you cannot map a given step to an action word found in the list, try to map synonyms of the instruction to an action word from the list. Either way, do not display any action words that are not in the action word list provided. Only assign one action word per step. Do not display headings for each step, only the action word. Here are the action words you can map to: [ACTION_LIST] Here is the text to process: [STEP_LIST]"* |
| Hallucination Reduction | Provide several additional prompt and a fixed action list to constrain outputs strictly to the allowed actions | Percentage of extracted actions that are members of the provided action list | 1) Provided the [ACTION_LIST] in the GPT prompt<br><br>2) Requested the GPT to only output given actions by adding the following to the prompt: *"These action words should only be from the action word list provided. If you cannot map a given step to an action word found in the list, try to map synonyms of the instruction to an action word from the list. Either way, do not display any action words that are not in the action word list provided."* |

| Activity Definition Refinement | Provide a list of synonyms and map each set to one canonical activity label; merge variants in the output and allow only that label | Percentage of extracted actions that are members of the provided action list | A python script checking for words with similar meanings and replace synonyms |

*3.3.2 Comparison of Skill Chaining Models*

In recent advancement of robot learning research, three popular models to imitate the sequence of elemental motions were reviewed. The following three models were chosen to perform the skill chaining task. Their model details are illustrated in the following sections.

1) Bayesian Network (BN) Learning (used Dynamic Bayesian Network (DBN), example: Tore et al. 2023).
2) Markov Decision Process (MDP)-based model (example: Suziki et al. 2007).
3) Language model: this paper will discuss LSTM (example: Zheng et al. 2024) and LLMs (GPT4o and GPT4 mini, example: Zhang et al. 2023).

The model size and training challenges are in ascending order. However, with the recent advancement of LLM, the efforts needed for the third category has largely been reduced.

The illustrations for these three models are shown below. Considering LSTM and LLM are most frequently used in construction robot learning papers, this section will provide details on BN and HMMBI, with less emphasis on LSTM and LLM.

Bayesian Network

BN learning is an effective method for representing structured knowledge and skills. A BN is structured as a network where various probabilistic nodes are connected by edges. It is essentially a Directed Acyclic Graph (DAG), as shown in Figure 9. The node in this DAG will be the event observed (can be the elemental action, the motion, or the command). The arcs, which are the arrows connecting the events/nodes, show the conditional dependencies of each event. In the case of robot learning, an event could be equivalent to the robot elemental motions/ micro skills. For example, for an installation task, Event 1 could refer to pick up, and Event 2 could be transport or handover.

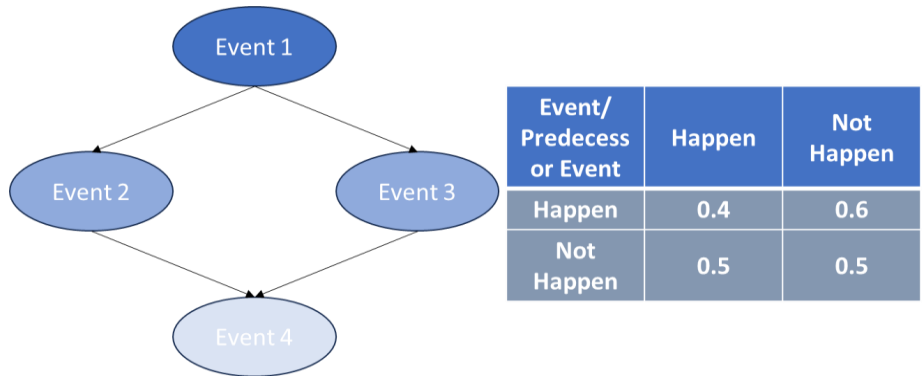

**Figure 9 An Example of Bayesian Network**

For example, if we use several categorical variables $X = \{X_1, X_2, X_3, X_4\}$ to represent the four events in Figure 8. For this DAG, it is essentially equivalent to a joint probability of:

$$p(\chi) = p(\chi_1, \chi_2, \chi_3, \chi_4) = \prod_i p(\chi_i | \pi_i) \quad Eq.(1)$$

In this equation,

$p(\chi)$ – the joint probability of all events;

$p(\chi_1, \chi_2, \chi_3, \chi_4)$ - the joint probability of event 1 to 4;

$p(\chi_i|\pi_i)$ - the conditional probability, meaning the probability of $\chi_i$ on that condition that its predecessor event $\pi_i$ happens;

The above equation represents the essence of Bayesian probability, which is to categorize the probability of one event in terms of both the original prior probability and the posterior probability that one event happens upon the observation of another event, as shown in the following theorem:

$$p(\chi_i|\chi_{i-1}) = \frac{p(\chi_i \cap \chi_{i-1})}{p(\chi_i)} = \frac{p(\chi_{i-1}|\chi_i) \times p(\chi_{i-1})}{p(\chi_i)} \quad Eq.(2)$$

The Structure Learning (SL) of BN was used to implement Eq. (2). SL refers to the problem of searching for the optimal next node and establishing the interdependency between the events and therefore gradually building the network (Scanagatta et al. 2019). It includes score-based leaning, constraint-based learning, and structural expectation–maximization (SEM) for incomplete data. The purpose of SL is to establish the tree network for the KN. In this paper, after examining two options, including Tree Augmented Naive Bayes (Friedman et al. 1997) and Chow-Liu (Chow and Liu 1968) algorithms, the SEM with Chow-Liu tree search algorithm was used to connect the elemental motions.

The Chow-Liu tree search algorithm is a greedy search algorithm that finds the most likely next node through the following four steps:

- trying to add node to span the tree. In this case, one node will be one action verb/ micro planning step.
- calculate the weight of the possible edge $I(X_i, X_j)$ corresponding to the added node with the following:

$$I(X_i, X_j) = \sum_{x_i, x_j} \hat{P}(x_i, x_j) \log \frac{\hat{P}(x_i, x_j)}{\hat{P}(x_i)\hat{P}(x_j)} \quad Eq.(3)$$

Where the practical distribution of $\hat{P}(x_i, x_j) = \frac{Count(x_i, x_j)}{Number\ of\ examples}$.

- calculate the total weight $\sum_{(X, X_j) \in E} I(X_i, X_j)$, where E is the set of all possible pairs of events/ action verbs.
- keep the option returning greatest total weight for every step.

Hidden Markov Model with Bayesian Inference

Denote the hidden states as $S = \{s_1, s_2, ..., s_n\}$ and observe sequence of events/ tokens $O = \{o_1, o_2, ..., o_n\}$. In this case, an event/ token refers to an elemental motion/ micro skill.

The transition probability, emission probability, and initial state probability are shown as $A, B, \pi$ respectively.

HMMBI aims to calculate the next state/ token by:

$$P(o_{t+1}|O) = \Sigma\, P(o_{t+1}|s_t) * P(s_t|O) \quad Eq.(4)$$

$$P(O|s_t) = \frac{\alpha(s_t) \times \beta(s_t)}{\Sigma \alpha(s) \times \beta(s)} \quad Eq.(5)$$

In this equation,

$\alpha(s_t) = P(o_{1:t}|s_t)$: Forward probability, the likelihood of observations up to t ending in state $s_t$.

$\beta(s_t) = P(o_{t+1:k+1}|s_t)$: Backward probability, the likelihood of future observations given state $s_t$.

3.3.3 Information Exchange for Executing Chained and Parameterized Skills

As mentioned in Section 3.1.2, humans can select the micro skills for robots to execute through a web application before the robot chaining process. With the above algorithms for the robot chaining system, the robot control and information flow will change correspondingly. The overview of the overall robot control scheme is adapted to a more comprehensive structure as shown in Figure 10. The system's responsiveness for rapid information exchange in real-time deployment was evaluated. Over the course of ten experimental trials, the system demonstrated an average response time of 0.067 seconds for transmitting semantic information from BIM to ROS. The high-speed data communication pipeline demonstrated its suitability for CRP.

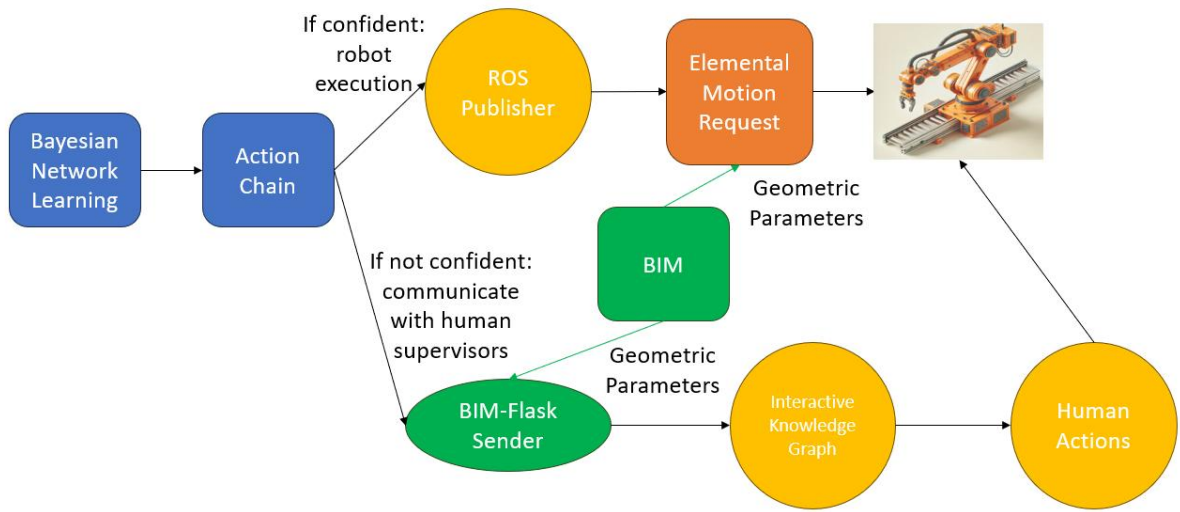

Figure 10 Information Flow for Parameterized and Chained Micro Skills

4. Experiments and Validation

Several experiments were conducted to validate the feasibility and benefits of the proposed framework. First, the continuity of the micro skill database was evaluated with manual inspections of the start and end state. Second, robot micro skill execution experiments were conducted to evaluate the comprehensiveness of the database. The third experiment is to compare the performances of the selected chaining algorithms. The authors created the next action prediction cases and examined the models' performances in predicting the next step actions. A summary of all experiments, validations, and their evaluation metrics are summarized in Table 4.

Table 4 Summary of Experiments and Metrics

| Experiments | Purpose | Metrics or Standards |
|---|---|---|
| Micro skill database examination | Validating the quality of proposed micro skill database | Continuity, so that a combination of all necessary words can represent all the object states the robot-held objects or tools needed to go through to complete a task, shown in Table 3 |
| Robot execution experiments | Examine that the proposed micro actions are robot system interpretable and executable and gap the language-robot action mismatch | Robot long-horizon task execution success, as shown in Figure 6 |
| BIM parametric parsing web-ROS system prototyping | Prototypes and validate the feasibility of the proposed micro skill parameterization system | Parameter parsing completeness and accuracy, by comparing original BIM information and ROS printed version; first line in Table 5 |
| Macro skill learning model suitability comparison | Experientially provide evidence to support the statement for the suitability and optimality for choosing LLM for hierarchical learning tasks | Accuracy of action sequencing, as shown in Figure 11 |

*4.1 Continuity and Comprehensiveness of the Micro Skill Knowledge Base*

The authors choose a drywall installation case to examine whether the proposed skill base can represent a randomly chosen construction task. The chaining process was conducted with the BN SL with Chow-Liu algorithm by implementing with Python and pgmpy library (considering its chaining quality was the easiest to inspect). The elemental words extracted from the installation tutorials and their associated object states are continuous, as shown in Table 5. Furthermore, as illustrated by the 'ObjectStateStart' and 'ObjectStateEnd' columns in Table 5, each node maintains the same end object state as the beginning object state of its predecessor node. The authors have assigned same colors for identical states to highlight the continuity of object state. This consistency confirms the continuity of the proposed method.

Table 5 Object State and Knowledge Graph Validity Analysis

| Name of Activity | ObjectStateBegin | ObjectStateEnd | Continuity and Validity |
|---|---|---|---|
| Start | | Coordinate same as material stack | Yes |
| Prepare | Coordinate same as material stack | Coordinate same as material stack | Yes |
| Plan | Coordinate same as material stack | Coordinate same as material stack | Yes |
| Cut | Coordinate same as material stack | Coordinate same as material stack, size becomes required size | Yes |
| Connect | Coordinate same as material stack, size becomes required size | Coordinate same as target, size becomes required size | Yes |
| Finish | Coordinate same as target, size becomes required size | | Yes |

*4.2 Robot Execution Experiment*

To ensure the validity, continuity, and comprehensiveness of the proposed workplan, a robot execution experiment was also performed. A human supervisor was asked to verify and approve the workflow by interacting with a web application to avoid potential effects from the errors of the chaining process. The supervisor confirmed the correct steps in the correct order by clicking buttons on the application (e.g., clicking "Plan" first, followed by "Prepare") to maximize the quality of skill chaining and safety in the experiment process. A screenshot of a sample webpage is shown in Figure 11. Upon clicking, a package called *flask_ROS* sends the command to ROS. A ROS subscriber listens for this signal, saves the approved and sequenced motions, and then allows the robot to plan and execute these motions in the correct sequence.

For each action step, a Python function was written. The subscriber calls each function in sequence, enabling the robot to execute the learned sequence of actions. It was observed that the transition of some micro-actions necessitated tool changes. This observation highlights the importance of close-proximity human-robot collaboration. The following solutions and rules are thus proposed: If the robot gripper is still in use but requires another tool for the upcoming task, the human worker will assist by performing the next task. If the robot gripper is no longer in use and a tool change is needed, the robot will alert the human worker, who will then perform the tool change.

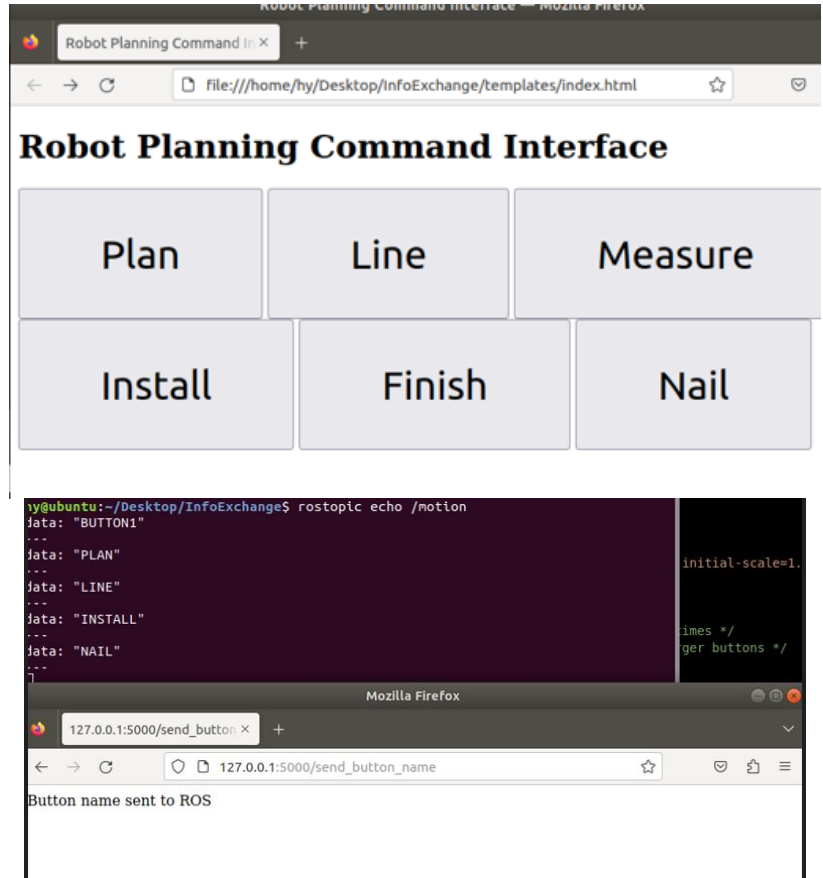

**Figure 11 Web Application for Human Commanding Robot and ROS Feedback**

The pseudocode and results are as shown in Table 6.

**Table 6 Micro Actions and Steps of Drywall Installation**

| Action | Definition and Process | Pseudocode |
|---|---|---|
| Plan | Retrieve information from BIM to localize target with rhino_ros package (Robot). Same information will be sent to web application through flask, rhino3dm, requests, json package. | **Task parameter retrieval and planning:**<br>*Rhino to ROS:*<br>  Configure connection with ROS;<br>  Retrieve object features and location;<br>  Send data;<br><br>*Rhino to web app:*<br>  Configure connection to web app server;<br>  Read BIM model using rhino3dm;<br>  Retrieve object features and location;<br>  Save object features and decode to transmissible format;<br>  Send data; |
| Prepare | Robot wait for human to check if the material is the correct one to use | **Human-robot collaboration:**<br>If workpiece correct:<br>  human confirm with robot<br>If workpiece incorrect:<br>  human request robot to move and pickup new objects |
| Cut | Robot plans a cutting trajectory and uses the cutting tool to cut the object | **Cutting trajectory planning:**<br>*Input*<br>*object center location in X-Y plane (x,y), object thickness z, cutting tool length L*<br>*original object dimension: length Lo and width Wo*<br>*required object dimension: length La and width Wa*<br><br>Move to start point:<br>  (x-Lo/2+La,y-Wo/2,z)<br>Insert and cut by moving to:<br>  (x-Lo/2+La,y-Wo/2,0)<br>Continue cutting and move to:<br>  (x-Lo/2+La,y-Wo/2+Wa,0)<br>Move to end point:<br>  (x-Lo/2,y-Wo/2+Wa,0)<br>Move away to :<br>  (x-Lo/2,y-Wo/2+Wa,L+4*z)<br>Stop<br>Request tool change |
| Install | Robot picks up the object, moves it to the target location, and places the object at the target place. Robot then alert human if needs help. | **Place object:**<br>Confirm target location<br>Plan a trajectory to target location<br>Execute trajectory<br>Hold panel and alert human |
| Nail | Robot wait for human to nail the panel. | Robot wait for human workers or another robot to nail |

## 5. Experimental Results

As mentioned in Section 4.1, the object state transformation continuity of the micro skills is confirmed. This subsection will focus on the robot execution results, as shown in Table 7.

Additionally, the performance of robot skill chaining was evaluated using a test case derived from training data that combined information from multiple domains. The results, presented in Figure 12, indicate that LLM (GPT-4o) and LSTM exhibited optimal performance in skill chaining. Beyond prediction accuracy, several additional metrics concerning model implementation and output were used to assess the suitability of Markov-based models compared to language-based models, as shown follows:

First, the level of difficulty in model parameter tuning: parameter tuning for LSTM proved significantly simpler than for the HMMBI method. Despite multiple rounds of adjustments to the hidden states, number of iterations, and initial transition probabilities, HMMBI failed to accurately predict the next action. Second, the data preprocessing efforts: HMMBI required data encoding without understanding synonyms, leading to higher demands on data preprocessing, synonym replacement, and micro skill definitions. Similar limitations were observed with DBN. While DBN performed well with smaller datasets (10-15 entries), it demanded flawless preprocessing, where words needed to have exactly the same meanings and has to be encoded carefully. Even after preprocessing with ChatGPT, time-consuming manual adjustments were necessary.

However, from the perspective of output readability and interpretability, DBN offers an advantage in its ability to directly output DAGs using Python network packages. In contrast, attempts to generate DAGs with GPT faced challenges integrating with network drawing packages. Moreover, the generalizability and potential to transfer learnt skills to a new domain is another metric: the authors tested LLM (GPT-4) with unseen data. Even without fine-tuning or Low-Rank Adaptation (LoRA), LLM demonstrated strong predictive and chaining capabilities for these tasks. This zero-shot inference capability is likely attributable to the extensive dataset used in GPT training (Zhang et al., 2024), which may include knowledge bases similar to the tutorials provided. Overall, these observations confirm that language models are more suitable for robot skill chaining tasks. The only limitations are the additional programming workload to combine the GPT model with the visualization models to simplify the validation process.

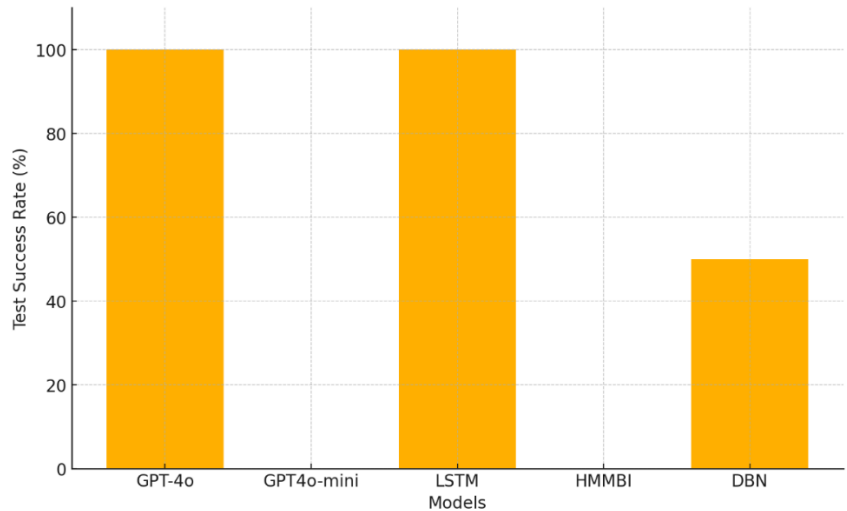

**Figure 12 Model Performance Comparison: GPT-4o-mini and HMMBI achieved a 0% success rate in the test, indicating unsuitability for zero-shot transfer learning under the given context.**

## 6. Discussion and Limitations

This paper provides a generalizable learning architecture for construction robots. Through computational experiments, the authors revealed the different level of suitability of sequence learning models for robot macro skill learning tasks. For example, GPT 4o can reach 100% accuracy while HMMBI cannot predict any next action correctly for unseen tasks with a prediction of accuracy of 0. Admittedly, there are several limitations:

1) This paper only experiments the pick-and-place based tasks for micro skills. There are two types of micro actions found during the review of online tutorial resources. They are both dependent on the geometric configurations in the task, including the object's dimension, location, and orientation. However, certain tasks have a higher dependency on the task parameters and carries diverse meaning. For example, the installation task of "drywall installation" refers to essentially a pick-and-place task. In contrast, the installation of "drop ceilings" requires more complicated manipulation of workpieces and thus requires different skills (Yu et al. 2024). For the second kind of actions, there are several papers in construction domain showing non-pick-and-place skills from workers to robots. They are also verified compatible with workers without programming expertise. The scope of this paper is not to transfer all construction skills to robots but to propose a continuous learning framework and knowledge base that are compatible with other learning frameworks. For example, elemental skills mentioned in ceiling installation papers can also be added to the micro skill base proposed in this paper and use the chaining models proposed to complete the learning process. The highlighted contribution from this paper is to have a generally applicable framework that enables continuous absorption and learning of other micro skills as robotics research advances.

2) Manual data handling processes and the need for more automation: Several critical aspects of our methodology are performed manually, including searching for tutorials, preprocessing (DBN and HMMBI model), inputting them into GPT, and inspecting ChatGPT's preprocessed results. These steps are essential to the process. However, there is a pressing need for more advanced and automated LLM pipelines to streamline the interpretation of these actions, to improve both efficiency and accuracy.

3) Single modality of data: This study focused exclusively on verbal instructions and tutorials in the training data. Our work is centered on the examination of natural language instructions and the role and effectiveness of LLMs in robot learning. Consequently, only language-based information was utilized. This choice is deemed as sufficient and justified in the Introduction. In future work, multi-modal training data, including video demonstrations, VR simulations, and verbal instructions, could be integrated. This approach would enable mapping the skill base to VR-collected coordinates or human-demonstrated trajectories in videos, allowing robots to more effectively imitate human skills and behaviors.

4) Effective visualization of the elemental motion chaining process: In this paper, the authors propose two robotic construction modes utilizing an elemental-motion-based robot programming method, also known as episodic programming. The first mode enables human operators to directly input the desired sequence of elemental motions by selecting them sequentially via a web application. This application integrates the selected motions with corresponding BIM data to automatically generate the necessary robot commands, as illustrated in Figure 9. The second mode allows the robot to be fully automatic by learning action sequences from crowdsourced verbal instructions, such as those found in YouTube videos and other online sources. This approach could benefit from an easier verification of extensive textual data method for human workers. To streamline and enhance the efficiency of this verification process, the authors will develop a Knowledge Graph (KG) to succinctly visualize step-by-step information. A preliminary user study confirmed the effectiveness of the KG in improving human-robot communication. Currently, the authors are conducting a larger-scale user study aimed at further validating the proposed approach's ability to enhance inspection accuracy in fully autonomous episodic robot programming.

5) In addition, this paper does not consider real-time action planning or inference. Our approach emphasizes task-level reasoning prior to execution, where the full sequence of actions (the skill chain) is determined at the outset using the micro skill chaining models such as an LLM. Once a subtask or an action is selected, the robot parses task-related information (e.g., from BIM) to identify the goal state, and the motion is executed using standard motion planning libraries. Our decision not to pursue real-time inference was also influenced by observations from several engineering and robotics system demonstrations, where real-time inference models often produced jittery or unstable trajectories. Such behavior, while technically responsive, risks damaging workpieces or reducing human trust in robotic systems, particularly in dynamic and unstructured construction environments. Therefore, we prioritized reliability and consistency in planned motion over reactive, real-time behavior.

## 7. Conclusions

This paper proposes a robot programming framework using hierarchical modeling, crowdsourced natural language instructions, and LLM, and makes the following technical contributions:

First, the research proposed a micro skill knowledge base that covers commonly used actions for assembly and construction tasks. The authors performed an extensive literature review to understand the names of the actions, their granularity, and meaning for robot skill learning papers. Observing that some actions are widely used but they have different meanings for different task domains and different human users, the authors standardize, parameterize, and modularize these actions as micro skills to be task-parametrizable with BIM and easily learnable from human natural language instructions.

Table 7 Robot Micro Skill Execution Results

| Action | Robot Execution Results |
|---|---|
| Plan | 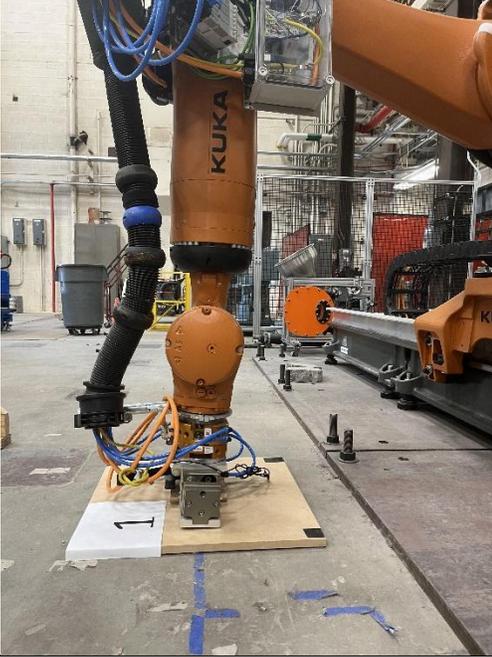<br>Screenshot of data sent from BIM to web applications |
| Prepare | Picture of robot moving to start location as the end state of "Prepare" |
| **Cut** | Picture of robot control interface signaling to the human supervisor that the cut process has started |
| **Install** | 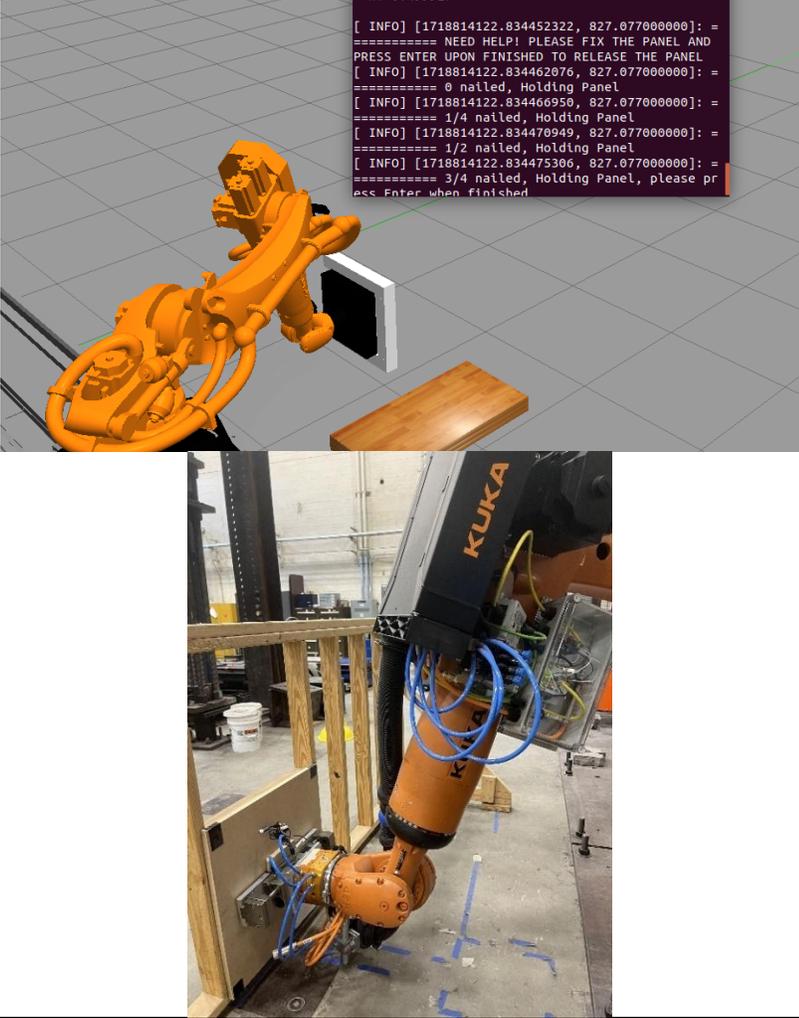 Pictures of robot reaching the end state for installation in simulation and in the real world |
| **Nail** | / |

Second, this work established a robot learning framework from crowdsourced natural language inputs and leverages LLM. With the above micro skill base, the robot learning problem for every domain is simplified to an action chaining problem. The challenge has shifted from reacting to every minor environmental observation to understanding how to effectively chain different micro-skills in the context of specific tasks. This approach to framing the learning problem mirrors and emulates human instructional and learning practices. To address this, the authors proposed an LLM-based framework for data retrieval, preprocessing, and analysis, designed to extract valuable action sequence information from crowdsourced and readily available inputs, including YouTube videos and professional construction tutorials. The proposed learning framework largely improved sample efficiency and reduced the human workload to program robots. It also has the potential to directly contribute to multiple task skill learning. For example, in a recent paper published in Nature, language models were used to establish the parallel robot learning scheme for efficacious multiple skill learning in a batch (van Diggelen et al. 2024).

Third, the authors experimentally tested computational models' suitability for the chaining process. The experiments compared the performance in skill chaining by running probability-based models (HMMBI and DBN) and language-based models (GPT 4o, GPT 4o mini, and LSTM). Through action prediction experiments, the authors reveal that language-based models have optimal performances for chaining micro skills extracted from natural language instructions.

Overall, this research innovates the robot learning framework by leveraging crowdsourced natural language instructions, LLM, and hierarchical modeling to improve the learning efficiency. This paper also examined the feasibility of learning skills from crowdsourced tutorials from YouTube and professional construction companies to minimize the workers' workload in programming robots. Its robot execution experiments verified the feasibility of using LLM to preprocess such inputs and select micro skills in a proper sequence to transfer the skills from human to robots.

**Data Availability Statement**

Some or all data, models, or code that support the findings of this study are available from the corresponding author upon reasonable request.

**Acknowledgements**

The authors would like to acknowledge the financial support for this research received from the US National Science Foundation (NSF) (Grant Nos. FW-HTF 2025805 and FW-HTF 2128623). Any opinions and findings in this paper are those of the authors and do not necessarily represent those of the NSF.

**References**

Aceves, P., & Evans, J. A. (2024). Human languages with greater information density have higher communication speed but lower conversation breadth. *Nature Human Behaviour, 8*(4), 644-656.

Agostini, A., Saveriano, M., Lee, D., & Piater, J. (2020). Manipulation planning using object-centered predicates and hierarchical decomposition of contextual actions. *IEEE Robotics and Automation Letters, 5*(4), 5629-5636.

Alias, R. A., & Rahman, A. A. (2016, August). User interface prototype through participatory cognitive task analysis. In *2016 4th International Conference on User Science and Engineering (i-USEr)* (pp. 105-110). IEEE.

Amani, M., & Akhavian, R. (2024). Adaptive Robot Perception in Construction Environments using 4D BIM. *arXiv preprint arXiv:2409.13837*.

An, Y., Yang, J., Li, J., He, B., Guo, Y., & Yang, G. Z. (2024, May). Skill learning in robot-assisted micro-manipulation through human demonstrations with attention guidance. In *2024 IEEE International Conference on Robotics and Automation (ICRA)* (pp. 15601-15607). IEEE.

Brosque, C., & Fischer, M. (2022). Safety, quality, schedule, and cost impacts of ten construction robots. *Construction robotics*, *6*(2), 163-186.

Cai, J., Du, A., Liang, X., & Li, S. (2023). Prediction-based path planning for safe and efficient human–robot collaboration in construction via deep reinforcement learning. *Journal of Computing in Civil Engineering*, *37*(1), 04022046.

Chen, J., Li, J., Huang, Y., Garrett, C., Sun, D., Fan, C., ... & Williams, B. C. (2022). Cooperative task and motion planning for multi-arm assembly systems. *arXiv preprint arXiv:2203.02475*.

Chen, Y., Ye, Y., Chen, Z., Zhang, C., & Ang, M. H. (2024). ARO: Large Language Model Supervised Robotics Text2Skill Autonomous Learning. *arXiv preprint arXiv:2403.15834*.

Denk, J., & Schmidt, G. (2001). Synthesis of a walking primitive database for a humanoid robot using optimal control techniques. In *Proceedings of IEEE-RAS international conference on humanoid robots* (pp. 319-326).

Ding, Y., Xu, W., Liu, Z., Zhou, Z., & Pham, D. T. (2019). Robotic task oriented knowledge graph for human-robot collaboration in disassembly. *Procedia CIRP, 83*, 105-110.

Ebert, F., Yang, Y., Schmeckpeper, K., Bucher, B., Georgakis, G., Daniilidis, K., ... & Levine, S. (2021). Bridge data: Boosting generalization of robotic skills with cross-domain datasets. *arXiv preprint arXiv:2109.13396*.

Escamilla, E., Ostadalimakhmalbaf, M., & Bigelow, B. F. (2016). Factors impacting Hispanic high school students and how to best reach them for the careers in the construction industry. *International Journal of Construction Education and Research, 12*(2), 82-98.

Feng, C., Fredricks, N., & Kamat, V. R. (2013). Human-robot integration for pose estimation and semi-autonomous navigation on unstructured construction sites. Ann Arbor, 1001, 48109.

Forte, D., Gams, A., Morimoto, J., & Ude, A. (2012). On-line motion synthesis and adaptation using a trajectory database. *Robotics and Autonomous Systems, 60*(10), 1327-1339.

Francis, A., Pérez-d'Arpino, C., Li, C., Xia, F., Alahi, A., Alami, R., ... & Martín-Martín, R. (2025). Principles and guidelines for evaluating social robot navigation algorithms. *ACM Transactions on Human-Robot Interaction*, *14*(2), 1-65.

Ha, H., Florence, P., & Song, S. (2023). Scaling up and distilling down: Language-guided robot skill acquisition. In *Conference on Robot Learning* (pp. 3766-3777). PMLR.

Hakhamaneshi, K., Zhao, R., Zhan, A., Abbeel, P., & Laskin, M. (2021). Hierarchical few-shot imitation with skill transition models. *arXiv preprint arXiv:2107.08981*.

Hönig, W., Ortiz-Haro, J., & Toussaint, M. (2022). db-A*: Discontinuity-bounded Search for Kinodynamic Mobile Robot Motion Planning. In *2022 IEEE/RSJ International Conference on Intelligent Robots and Systems (IROS)* (pp. 13540-13547). IEEE.

Huang, L., Zhu, Z., & Zou, Z. (2023). To imitate or not to imitate: Boosting reinforcement learning-based construction robotic control for long-horizon tasks using virtual demonstrations. *Automation in Construction*, *146*, 104691.

Intelligence, P., Black, K., Brown, N., Darpinian, J., Dhabalia, K., Driess, D., ... & Zhilinsky, U. (2025). $\pi_{0.5}$: a Vision-Language-Action Model with Open-World Generalization. *arXiv preprint arXiv:2504.16054*.

Janidarmian, M., Roshan Fekr, A., Radecka, K., Zilic, Z., & Ross, L. (2015). Analysis of motion patterns for recognition of human activities. In *Proceedings of the 5th eai international conference on wireless mobile communication and healthcare* (pp. 68-72).

Jiang, C., & Jagersand, M. (2020). Bridging visual perception with contextual semantics for understanding robot manipulation tasks. In *2020 IEEE 16th International Conference on Automation Science and Engineering (CASE)* (pp. 1447-1452). IEEE.

Jiang, M., Sun, D., Li, Q., & Wang, D. (2020). The usability of ventilator maintenance user interface: a comparative evaluation of user task performance, workload, and user experience. *Science progress*, *103*(4), 0036850420962885.

Jiang, R., He, B., Wang, Z., Cheng, X., Sang, H., & Zhou, Y. (2024). Robot skill learning and the data dilemma it faces: a systematic review. *Robotic Intelligence and Automation, 44*(2), 270-286.

Johannsmeier, L., Gerchow, M., & Haddadin, S. (2019). A framework for robot manipulation: Skill formalism, meta learning and adaptive control. In *2019 International Conference on Robotics and Automation (ICRA)* (pp. 5844-5850). IEEE.

Kazaz, A., Ulubeyli, S., & Tuncbilekli, N. A. (2012). Causes of delays in construction projects in Turkey. *Journal of Civil Engineering and Management, 18*(3), 426-435.

Kim, S., Chang, S., & Castro-Lacouture, D. (2020). Dynamic modeling for analyzing impacts of skilled labor shortage on construction project management. *Journal of Management in Engineering, 36*(1), 04019035.

Lee, R. K. J., Zheng, H., & Lu, Y. (2024). Human-robot shared assembly taxonomy: A step toward seamless human-robot knowledge transfer. *Robotics and Computer-Integrated Manufacturing, 86*, 102686.

Li, S., Fan, J., Zheng, P., & Wang, L. (2021). Transfer learning-enabled action recognition for human-robot collaborative assembly. *Procedia CIRP*, *104*, 1795-1800.

Li, Y., Wen, H., Wang, W., Li, X., Yuan, Y., Liu, G., ... & Liu, Y. (2024). Personal llm agents: Insights and survey about the capability, efficiency and security. *arXiv preprint arXiv:2401.05459*.

Liu, C., Cao, X., Xue, A., & Li, X. (2022). Motion Languages for Robot Manipulation. In *International Conference on Cognitive Computation and Systems* (pp. 299-314).

Lu, K., Ly, K. T., Hebberd, W., Zhou, K., Havoutis, I., & Markham, A. Learning Generalizable Manipulation Policy with Adapter-Based Parameter Fine-Tuning.

Mayr, M. (2024). Learning with skill-based robot systems: Combining planning & knowledge representation with reinforcement learning.

Miao, S., Zhong, D., Miao, R., Sun, F., Wen, Z., Huang, H., ... & Wang, N. (2022). Hierarchical Knowledge Representation of Complex Tasks Based on Dynamic Motion Primitives. In *International Conference on Cognitive Systems and Signal Processing* (pp. 452-462).

Nah, M. C., Lachner, J., & Hogan, N. (2024). Robot control based on motor primitives: A comparison of two approaches. *The International Journal of Robotics Research*, *43*(12), 1959-1991.

Nasiriany, S., Maddukuri, A., Zhang, L., Parikh, A., Lo, A., Joshi, A., ... & Zhu, Y. (2024). RoboCasa: Large-Scale Simulation of Everyday Tasks for Generalist Robots. *arXiv preprint arXiv:2406.02523*.

Niekum, S., Osentoski, S., Konidaris, G., Chitta, S., Marthi, B., & Barto, A. G. (2015). Learning grounded finite-state representations from unstructured demonstrations. *The International Journal of Robotics Research, 34*(2), 131-157.

Nozaki, T., Mizoguchi, T., & Ohnishi, K. (2014). Motion expression by elemental separation of haptic information. *IEEE Transactions on Industrial Electronics, 61*(11), 6192-6201.

Ortiz-Haro, J., Hönig, W., Hartmann, V. N., Toussaint, M., & Righetti, L. (2024). iDb-RRT: Sampling-based Kinodynamic Motion Planning with Motion Primitives and Trajectory Optimization. *arXiv preprint arXiv:2403.10745*.

Paulheim, H. (2017). Knowledge graph refinement: A survey of approaches and evaluation methods. *Semantic web, 8*(3), 489-508.

Paulius, D., Agostini, A., & Lee, D. (2023). Long-Horizon Planning and Execution with Functional Object-Oriented Networks. *IEEE Robotics and Automation Letters*.

Pedersen, M. R., Nalpantidis, L., Andersen, R. S., Schou, C., Bøgh, S., Krüger, V., & Madsen, O. (2016). Robot skills for manufacturing: From concept to industrial deployment. *Robotics and Computer-Integrated Manufacturing, 37*, 282-291.

Peng, S., Hu, X., Yi, Q., Zhang, R., Guo, J., Huang, D., ... & Li, L. (2023). Self-driven grounding: Large language model agents with automatical language-aligned skill learning. *arXiv preprint arXiv:2309.01352*.

Preece, J., Rogers, Y., Sharp, H., Benyon, D., Holland, S., & Carey, T. (1994). *Human-computer interaction*. Addison-Wesley Longman Ltd..

Redis, A. C., Sani, M. F., Zarrin, B., & Burattin, A. (2024). Skill Learning Using Process Mining for Large Language Model Plan Generation. *arXiv preprint arXiv:2410.12870*.

Saxena, A., Jain, A., Sener, O., Jami, A., Misra, D. K., & Koppula, H. S. (2014). Robobrain: Large-scale knowledge engine for robots. *arXiv preprint arXiv:1412.0691*.

Scanagatta, M., Salmerón, A., & Stella, F. (2019). A survey on Bayesian network structure learning from data. *Progress in Artificial Intelligence, 8*(4), 425-439.

Sener, F., Chatterjee, D., Shelepov, D., He, K., Singhania, D., Wang, R., & Yao, A. (2022). Assembly101: A large-scale multi-view video dataset for understanding procedural activities. In *Proceedings of the IEEE/CVF Conference on Computer Vision and Pattern Recognition* (pp. 21096-21106).

Sheils, J. (1988). *Communication in the modern languages classroom* (No. 12). Council of Europe.

Stenmark, M., & Malec, J. (2015). Knowledge-based instruction of manipulation tasks for industrial robotics. *Robotics and Computer-Integrated Manufacturing, 33*, 56-67.

Stewart, L. (2022). *Business Strategies for Reducing Professional Labor Shortages Within the US Construction Industry* (Doctoral dissertation, Walden University).

Suzuki, N., Hirasawa, K., Tanaka, K., Kobayashi, Y., Sato, Y., & Fujino, Y. (2007). Learning motion patterns and anomaly detection by human trajectory analysis. In *2007 IEEE International Conference on Systems, Man and Cybernetics* (pp. 498-503). IEEE.

Sweis, G., Sweis, R., Hammad, A. A., & Shboul, A. (2008). Delays in construction projects: The case of Jordan. *International Journal of Project Management, 26*(6), 665-674.

Team, O. M., Ghosh, D., Walke, H., Pertsch, K., Black, K., Mees, O., ... & Levine, S. (2024). Octo: An open-source generalist robot policy. *arXiv preprint arXiv:2405.12213*.

Tenorth, M., & Beetz, M. (2017). Representations for robot knowledge in the KnowRob framework. *Artificial Intelligence, 247*, 151-169.

Tenorth, M., Perzylo, A. C., Lafrenz, R., & Beetz, M. (2012). The roboearth language: Representing and exchanging knowledge about actions, objects, and environments. In *2012 IEEE International Conference on Robotics and Automation* (pp. 1284-1289). IEEE.

Tore, O. B., Negahbani, F., & Akgun, B. (2023). Keyframe demonstration seeded and bayesian optimized policy search. *arXiv preprint arXiv:2301.08184*.

van Diggelen, F., Cambier, N., Ferrante, E., & Eiben, A. E. (2024). A model-free method to learn multiple skills in parallel on modular robots. *Nature Communications, 15*(1), 6267.

Wang, X., Liang, C. J., Menassa, C. C., & Kamat, V. R. (2021). Interactive and immersive process-level digital twin for collaborative human–robot construction work. *Journal of Computing in Civil Engineering, 35*(6), 04021023.
50
Under review for ASCE OPEN: Multidisciplinary Journal of Civil Engineering

Wang, X., Wang, S., Menassa, C. C., Kamat, V. R., & McGee, W. (2023). Automatic high-level motion sequencing methods for enabling multi-tasking construction robots. *Automation in Construction*, *155*, 105071.

Ye, Y., You, H., & Du, J. (2023). Improved trust in human-robot collaboration with ChatGPT. *IEEE Access*, *11*, 55748-55754.

Yu, H., Kamat, V. R., Menassa, C. C., McGee, W., Guo, Y., & Lee, H. (2023). Mutual physical state-aware object handover in full-contact collaborative human-robot construction work. *Automation in Construction*, *150*, 104829.

Yu, H., Kamat, V. R., Menassa, C. C., McGee, W., Guo, Y., & Lee, H. (2023). Grip state recognition for enabling safe human-robot object handover in physically collaborative construction work. In *Computing in civil engineering 2023* (pp. 787-795).

Yu, H., Kamat, V. R., & Menassa, C. C. (2024). Cloud-based hierarchical imitation learning for scalable transfer of construction skills from human workers to assisting robots. *Journal of Computing in Civil Engineering*, *38*(4), 04024019.

Zhang, J., Wang, P., & Gao, R. X. (2021). Hybrid machine learning for human action recognition and prediction in assembly. *Robotics and Computer-Integrated Manufacturing, 72*, 102184.

Zhang, J., Zhang, J., Pertsch, K., Liu, Z., Ren, X., Chang, M., ... & Lim, J. J. (2023). Bootstrap your own skills: Learning to solve new tasks with large language model guidance. *arXiv preprint arXiv:2310.10021*.

Zhang, J., Ullah, N., & Babbar, R. (2024). Zero-Shot Learning Over Large Output Spaces: Utilizing Indirect Knowledge Extraction from Large Language Models. *arXiv preprint arXiv:2406.09288*.

Zheng, B., Verma, S., Zhou, J., Tsang, I. W., & Chen, F. (2022). Imitation learning: Progress, taxonomies and challenges. *IEEE Transactions on Neural Networks and Learning Systems*, *35*(5), 6322-6337.

Zheng, D., Yan, J., Xue, T., & Liu, Y. (2024). A knowledge-based task planning approach for robot multi-task manipulation. *Complex & Intelligent Systems, 10*(1), 193-206.

# Appendix 1 Mentioned Action Verbs

Table 1 Full list of action verbs for micro skills

| Column 1 | Column 2 | Column 3 | Column 4 |
| --- | --- | --- | --- |
| Acquire | Adjust | Adjust Position | Affix |
| Align | Alter | Apply Pressure | Approach |
| Arrange | Ascend | Assemble | Attach |
| Avoid | Bend | Bolt | Build |
| Calibrate | Carry | Carry Out | Categorize |
| Change | Choose | Clamp | Clear |
| Cling To | Close | Clutch | Collect |
| Come Closer | Compress | Connect | Construct |
| Convert | Convey | Coordinate | Create |
| Cut | Descend | Detach | Diagonal-Move |
| Direct | Disconnect | Disengage | Drag |
| Draw | Draw In | Draw Near | Drift |
| Drill | Elevate | Eliminate | Embed |
| Evade | Execute | Explore | Extract |
| Fasten | Fill | Find | Fine-Tune |
| Fix | Fold | Follow | Free |
| Grab | Grasp | Grip | Guide |
| Handover | Haul | Hold | Hold In Place |
| Horizontal-Move | Identify | Implement | Insert |
| Install | Investigate | Lay | Let Go |
| Level | Lift | Load | Locate |
| Look For | Maneuver | Mark | Measure |
| Modify | Monitor | Mount | Move |
| Move Back | Move Relative | Move Towards | Navigate |
| Obtain | On Hold | Open | Order |
| Organize | Oscillate | Pack | Pass |
| Patch | Perform | Pick | Pick Up |
| Pilot | Pinch | Pinpoint | Place |
| Place Down | Place Up | Plan | Pluck |
| Position | Press | Propel | Pull |
| Pursue | Push | Put | Put Down |
| Put Together | Raise | Rearrange | Recede |
| Release | Relinquish | Relocate | Remain Ready |
| Remove | Reorder | Reorganize | Reposition |
| Reshuffle | Retreat | Retrieve | Rotate |
| Screw | Seal | Search | Secure |
| Seek | Seize | Select | Separate |
| Set | Set Down | Set Up | Shake |
| Shift | Shut | Slide | Slot In |

55| Sort | Spin | Squeeze | Standby |
|------|------|---------|---------|
| Steer | Steer Clear Of | Take | Take Away |
| Thrust | Tighten | Track | Transfer |
| Transition | Transport | Traverse Diagonally | Traverse Horizontally |
| Traverse Vertically | Trigger | Turn | Tweak |
| Twist | Unfasten | Unlock | Vertical-Move |
| Wait | Wander | Withdraw | |

55
**Under review for ASCE OPEN: Multidisciplinary Journal of Civil Engineering**